\documentclass{article} 
\usepackage{iclr2025_conference,times}
\usepackage{float}
\usepackage{graphicx}

\usepackage{amsmath,amsfonts,bm}

\def\eqref#1{equation~\ref{#1}}

\def\1{\bm{1}}

\DeclareMathAlphabet{\mathsfit}{\encodingdefault}{\sfdefault}{m}{sl}
\SetMathAlphabet{\mathsfit}{bold}{\encodingdefault}{\sfdefault}{bx}{n}

\usepackage{hyperref}
\usepackage{url}

\title{Stacking Small Language Models for Generalizability}

\author{Laurence Liang \thanks{All correspondences can be sent via email to laurence.liang [at] mail.mcgill.ca } \\
McGill University\\
}

\iclrfinalcopy 
\begin{document}

\maketitle

\begin{abstract}
Recent advances show that large language models (LLMs) generalize strong performance across different natural language benchmarks. However, the large size of LLMs makes training and inference expensive and impractical to run in resource-limited settings. This paper introduces a new approach called fine-tuning stacks of language models (FSLM), which involves stacking small language models (SLM) as an alternative to LLMs. By fine-tuning each SLM to perform a specific task, this approach breaks down high level reasoning into multiple lower-level steps that specific SLMs are responsible for. As a result, FSLM allows for lower training and inference costs, and also improves model interpretability as each SLM communicates with the subsequent one through natural language. By evaluating FSLM on common natural language benchmarks, this paper highlights promising early results toward generalizable performance using FSLM as a cost-effective alternative to LLMs.\end{abstract}

\section{Introduction}

Since the publication of the transformer paper \cite{vaswani2017attention}, a considerable amount of research devoted to large language models (LLMs) has shown that LLMs are capable of generalizing well on natural language benchmarks and that new emergent properties appear as LLMs increase in scale. \cite{devlin2019bertpretrainingdeepbidirectional, wei2022emergentabilitieslargelanguage}. LLMs seem to follow some empirical scaling laws, where larger datasets, compute and model size contribute to improvements in model performance. \cite{kaplan2020scalinglawsneurallanguage}

As language models and datasets increase in size, a growing need emerges to identify methods to run language models in resource-limited settings where large amounts of compute are inaccessible. In fact, multiple methods have been documented and researched in recent years to make LLM training or inference more computationally efficient. One such method is fine-tuning: given a pre-trained model, fine-tuning that model for specific tasks can cause that model to score better on benchmarked tasks downstream. \cite{brown2020languagemodelsfewshotlearners} Furthermore, more efficient methods of fine-tuning such as LoRA and QLoRA also show that adding a trainable adapter to LLMs whose weights are frozen also allows for faster fine-tuning while showing strong signs of solid model performance. \cite{hu2021loralowrankadaptationlarge, dettmers2023qloraefficientfinetuningquantized}

Additionally, recent work indicates that small language models (SLM), such as Microsoft's Phi-3, can still achieve decent performance on natural language benchmarks. This finding is important, as it suggest that small language models, which are a few orders of magnitude smaller than state-of-the-art LLMs, can still achieve solid performance on various benchmarks. \cite{abdin2024phi3technicalreporthighly}

This paper aims to build on both the fine-tuning and small language model directions, in order to identify methods that allow for cost-effective training and inference in resource-limited settings. As a result, this paper proposes a new model framework called Fine-tuning Stacks of Language Models (FSLM) - or "stacking" - which involves chaining multiple specialized small language models together such that the framework's input and output resemble those of performant language models. 

FSLM takes loose inspiration from the human brain, where different components specialize in different tasks. For small language models, because each SLM has limited capabilities due to its small size, FLSM aims to fine-tune each SLM to specialize in a specific task. As a result, the motivating question becomes: how small can the SLMs be, such that the fine-tuned stack of SLMs is still capable of generalizing on various natural language benchmarks?

Our work challenges the lower-bound for SLM size by evaluating an FSLM stack of four Pythia models of 160 million parameters each. \cite{biderman2023pythiasuiteanalyzinglarge} By fine-tuning this FSLM stack on the Alpaca dataset, and benchmarking FSLM and models of similar size, this paper shows that FSLM stacks show promise as lightweight alternatives to heavier LLMs. 

Thus, this paper's contributions can be summarized as:

\begin{itemize}
    \item Proposing the FSLM stack as a lightweight framework to evaluate small language models in resource-limited settings.
    \item Introducing model distillation to fine-tune SLMs in order to minimize the need for human supervision or labeling. 
    \item Identifying early signs of FSLM generalizability by comparing FSLM of Pythia-160M models with Pythia and Flan models of comparable sizes
    \item Documenting model explainability by looking at the intermediary outputs between SLMs in the FSLM stack.  
\end{itemize}

\section{Related Work}

\subsection{Model Fine-Tuning}

In recent years, researchers have shown that pre-training a language model in a self-supervised fashion, followed by fine-tuning that same model to a variety of tasks, improves model performance downstream on natural language benchmarks. OpenAI's GPT is a notable example of fine-tuning a pre-trained model.  \cite{brown2020languagemodelsfewshotlearners} Because fine-tuning entire models is expensive, researchers have developed different methods to minimize computational cost while still achieving similar model performance.

\cite{hu2021loralowrankadaptationlarge} introduced \textbf{Low-Rank Adaptation (LoRA)} as a fine-tuning approach. LoRA freezes the weights of the original pre-trained model, and adds an "adapter" component, located between the original model output and the actual text output. Instead of the adapter being a fully connected layer, the adapter uses matrix factorization to generate low-rank matrix multiplications that approximate the fully connected equivalent. Low-rank matrix multiplication, however, is less computationally expensive than running inference on a fully connected layer. \cite{hu2021loralowrankadaptationlarge} then show that LoRA can maintain or even improve model performance. \cite{dettmers2023qloraefficientfinetuningquantized} developed \textbf{QLoRA}, which performs quantization to further improve LoRA. Both QLoRA and LoRA are considered to be \textbf{Parameter-Efficient Fine-Tuning (PEFT)} methods, a group of methods that aim to increase the efficiency of fine-tuning models. \cite{xu2023parameterefficientfinetuningmethodspretrained}

\subsection{Model Compression}

Model compression techniques aim to either shrink a given model's size, or to train a smaller model to learn from a larger one. 

For instance, \textbf{quantization} reduces the precision of the model weights, thus decreasing the overall size of the model. Even though the model loses precision, if quantization is implemented correctly, the model should maintain a similar level of performance while experiencing a speedup for training and inference. \cite{jacob2017quantizationtrainingneuralnetworks}

\textbf{Model pruning} removes weights whose values are close to zero, thus eliminating weights that may not be contributing to the model's main inference. \cite{cheng2024surveydeepneuralnetwork}

\textbf{Model distillation} is another method of interest: using a teacher-student architecture, a smaller "student" model learns from a larger "teacher" model that should be already well-trained. As a result, the teacher model distills its internal knowledge to the student model, by providing the student model inputs and outputs to learn from during this training process. \cite{hinton2015distillingknowledgeneuralnetwork, sanh2020distilbertdistilledversionbert}


\section{Method}

\begin{figure}[H]
    \centering
    \includegraphics[width=0.8\linewidth]{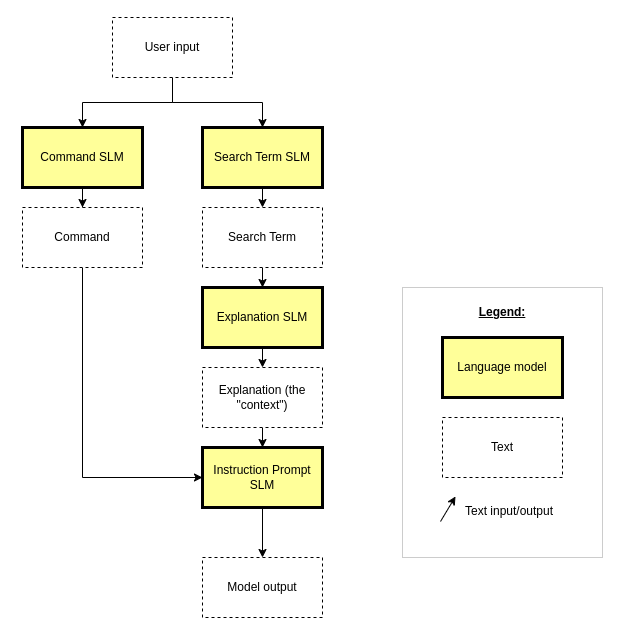}
    \caption{A visual representation of the FSLM stack. }
    \label{fig:fslm_flow}
\end{figure}

\subsection{FSLM Method Overview}

The FSLM framework consists of four small language models (SLM) that each specialize in a specific task, as shown in Fig. \ref{fig:fslm_flow}. A human user would supply a prompt to the FSLM framework, and the FSLM framework responds with a textual output. Internally, the SLMs look for specific textual elements from either the user's input or another SLM's output. As a result, each individual SLM is compensating for its limited capabilities by instead specializing in a specific task. As a result, the overall framework follows an information flow where textual information is slowly processed towards the intended model output. 

\subsection{Choice of Models}

We use the Pythia 160M GPT-NeoX architecture from the Pythia suite, as Pythia allows for ease of future scalability as we can evaluate on different model sizes. \cite{biderman2023pythiasuiteanalyzinglarge} Pythia also integrates well with LM-eval, which we use to evaluate FSLM on natural language benchmarks. \cite{gao2024eval-harness} 

\subsection{Choice of Dataset}

We use the Alpaca dataset to train FSLM in an instruction-tuning manner. \cite{alpaca} Alpaca contains 52,000 self-instruct generated instructions covering a wide array of applications. As of this writing, we selected a subsample of 5,000 instructions to fine-tune FSLM. 

\subsection{Training Data Generation}

In order to properly distill the intermediary texts between SLMs, we use the Llama 3.2 (3B) model to generate texts, a recent addition to the Llama family of LLMs. \cite{touvron2023llamaopenefficientfoundation}

\subsection{Fine-Tuning}

We use HuggingFace's PEFT implementation to run LoRA for fine-tuning. 

\section{Experiments}

\subsection{Natural Language Benchmarks}

We use Eleuther AI's LM-Evaluation Harness to run natural language tasks from TinyBenchmarks. \cite{gao2024eval-harness, polo2024tinybenchmarksevaluatingllmsfewer}

\begin{table}[H]
    \centering
    \begin{tabular}{l|cc}
        Model & tinyArc & tinyMMLU \\
     \hline\hline 
        FSLM (4x Pythia-160M) & {0.3349} & {0.3208} \\

        \hline
        Pythia-160M (no adapter) & 0.3213 & 0.3014 \\
        Pythia-1B (no adapter) & 0.2945 & 0.2720 \\

        \hline
        Flan-T5-Base (250M) (no adapter) & 0.2781 & 0.3615 \\
        Flan-T5-Large (780M) (no adapter) & 0.4209 & 0.4415 \\

    \end{tabular}
    \caption{Natural language benchmark results. All tasks are zero-shot, accuracy is the scoring metric. All Pythia models are taken from step 130,000.}
    \label{tab:zero_shot_nlu_tasks}
\end{table}

From Table \ref{tab:zero_shot_nlu_tasks}, we observe that our FSLM stack (following fine-tuning) performs better than non-adapter 160M and 1B Pythia models on tinyArc and tinyMMLU. This shows that fine-tuning specialized models in a "stack" does not worsen overall model performance compared to vanilla Pythia models of comparable size - rather, FSLM actually observes an increase in performance relative to Pythia models. 

Even though our FSLM implementation performs better than Google's Flan-T5-Base on tinyArc, Flan-T5-Base's performance on tinyMMLU is higher than FSLM's. Notably, Flan-T5-Large outperforms FSLM on both tasks by a noticeable margin. While FSLM on Pythia-160M shows encouraging early signs, it's possible that Flan-T5 models may exhibit superior performance due to their pre-training or fine-tuning processes. \cite{chung2022scalinginstructionfinetunedlanguagemodels} As a result, it becomes relevant to use different pre-trained models for FSLM and to run fine-tuning on different datasets as future experiments to implement.

\subsection{Qualitative Analysis for Model Interpretability}

Our FSLM implementation with four Pythia-160M is capable of simple question and answering in a coherent manner, as shown in Table \ref{tab:qualitative_sample}. Since our FSLM framework (approximately 640M parameters, or around 1.4 GB) is sufficiently small to run on most mobile phones or personal computers, the coherence and natural-sounding response further show promise that FSLM can run in resource-limited settings and exhibit human-like responses.

The intermediary outputs of SLMs within FSLM is of particular interest, because these responses allow us to directly evaluate model interpretability. Accordingly, we observe in Table \ref{tab:qualitative_sample} that the intermediary SLM outputs match very strongly with each pre-defined task, at least from the perspective of a human observer. While this shows that LoRA fine-tuning for FSLM is cost-effective, these intermediate SLM responses also serve as a checkpoint to flag potential mistakes or hallucinations. Because each SLM is specialized for a specific task, we expect the scope of the responses for each SLM to be somewhat bounded. As a result, if we detect that one of the responses seems wrong, it may be sufficient to only re-tune that single SLM, instead of the whole FSLM stack. In addition to promoting model explainability, this design would also minimize compute costs needed to fix overall model performance throughout model deployment. 

\begin{table}[H]
    \centering
    \begin{tabular}{p{2.1cm}p{5cm}}
        Section & Text \\
        \hline \hline
        
        User input & \textbf{What are ways to stay healthy?} 
        \\
        \hline \hline 
        Instruction & Find a healthy diet. \\ \hline
        Search term & "How to keep your body healthy" \\ \hline
        Explanation (the "context") & The most important thing to remember is to keep your  body healthy. If you don't, your body will be damaged and you'll be unable to perform the essential tasks of daily living. \\        
        \hline \hline
        \\
        Final output & \textbf{Healthy eating is a key component of maintaining a healthy lifestyle.} 
    \end{tabular}
    \caption{Sample breakdown of the intermediary texts generated by FSLM. }
    \label{tab:qualitative_sample}
\end{table}

However, throughout our model development process, we observed that FSLM responses can vary from one inference call to the next. As a result, future work should investigate optimal model temperature and top-k and top-p values in order to ensure repeatability and minimize high variances in model responses.

\section{Conclusion and Discussion}

The objective of this paper was to evaluate whether FSLM, a stack of task-specific SLMs, can perform well on natural language benchmarks and also exhibit natural-sounding text responses. By running natural language benchmarks, we determined that there were promising signs showing that FSLM's Pythia models perform on par with vanilla Pythia models of comparable sizes, suggesting that stacking fine-tuned specialized models can lead to accurate models at small scales. Additionally, by observing the full response of a sample model output, we determined that the final output was coherent and natural-sounding, and that the intermediary outputs were also highly aligned to each SLM's intended task. Additionally, FSLM's modular design could allow for easy model debugging and replacement of faulty SLMs. These results demonstrate encouraging signs that stacks of highly specialized small language models can perform as well as equivalent models of the same size, making FSLM architectures a potential area of interest for resource-limited compute settings. 

One main limitation concerns the limited scope for natural language benchmark evaluations. Because FSLM is a new implementation, we needed to write additional code to integrate it with existing lm-eval tasks, which initially limited the scope of tasks we could run as of this writing. Consequently, future work should increase the number of natural language benchmarks, and also evaluate model perplexity for token generation, and rouge scores for model summarization. Furthermore, surveys with human observers interacting with FSLM would be beneficial, as we would be able to quantitatively assess the quality and helpfulness of human-to-model interactions. 

Another limiting factor is the fine-tuning scope. Future work should try different fine-tuning datasets and determine to what extent dataset quality influences model performance downstream. On a similar topic, model pre-training should also be documented, as shown by the flan-T5 models' superior performances. Future work should investigate fine-tuning SLMs across different architectures that underwent different pre-training processes.

\section{Reproducibility Statement}

All the code used in this paper is accessible publicly on GitHub. The code is written in Jupyter Notebooks, which makes it easy for researchers to run and reproduce these results. Due to the double-blind submission, the GitHub link is not displayed here, though the codebase is available upon request. 

\bibliography{iclr2025_conference}
\bibliographystyle{iclr2025_conference}


\end{document}